%% file: main.tex
\documentclass[sigconf]{acmart}
\usepackage{algorithm}
\usepackage{algpseudocode}
\usepackage{amsmath}
\usepackage{graphicx}
\usepackage{xcolor}
\usepackage{subcaption}
\usepackage[normalem]{ulem}
\usepackage{tabularx}
\usepackage{array}

\providecommand{\bl}[1]{#1}
\providecommand{\rev}[1]{#1}

\AtBeginDocument{}

\setcopyright{none}
\acmConference[WWW]{The Web Conference}{}{}
\acmYear{}
\copyrightyear{}
\acmDOI{}
\acmISBN{}
\title[DGC-Net]{DGC-Net: Cross-Modal Dynamic Routing with Global-Guided Compensation in Multimodal Sentiment Analysis}

\author{Linan ZHU}
\affiliation{%
  \department{College of Computer Science and Technology}
  \institution{Zhejiang University of Technology}
  \city{Hangzhou}
  \country{China}
}

\author{Wu Huaijing}
\affiliation{%
  \department{College of Computer Science and Technology}
  \institution{Zhejiang University of Technology}
  \city{Hangzhou}
  \country{China}
}

\author{Xiao Han}
\affiliation{%
  \department{College of Computer Science and Technology}
  \institution{Zhejiang University of Technology}
  \city{Hangzhou}
  \country{China}
}

\author{Yuhao Wang}
\affiliation{%
  \institution{AIDC, Alibaba Group}
  \country{China}
}
\affiliation{%
  \department{Department of Data Science}
  \institution{City University of Hong Kong}
  \city{Hong Kong SAR}
  \country{China}
}

\renewcommand{\shortauthors}{Zhu et al.}

\begin{document}

% Avoid stretching heading/paragraph glue to force equal column bottoms.
\raggedbottom
% Allow the line-breaking algorithm a fallback for long technical phrases.
\setlength{\emergencystretch}{3em}

\begin{abstract}

\bl{User-generated videos, including opinion vlogs and
video reviews, provide a rich source of people's attitudes toward products, services, and social topics. Understanding
the sentiment expressed by speakers is important for analyzing online opinions, as transcripts alone may not capture the attitude and intensity conveyed through vocal tone and facial expressions. Multimodal sentiment analysis (MSA) addresses this need by integrating textual, acoustic, and visual cues. However, these cues are often subtle, noisy, and inconsistent across modalities.} Existing local affective cue modeling methods typically operate within individual modalities and overlook global contexts from other modalities. This may cause cross-modal affective cue inconsistency. Complementarily, hierarchical global--local fusion methods integrate global semantics with fine-grained local affective cues. However, they commonly determine the relative contributions of the two representations implicitly, preventing the global--local contributions from adapting to the sample-dependent reliability of local cues and resulting in either insufficient use of informative local cues or excessive interference from unreliable ones. To address these issues, we propose DGC-Net, a Dynamic Gating and Global-Guided Compensation Network for MSA. DGC-Net mitigates cross-modal affective cue inconsistency through cross-modal contribution-aware dynamic expert routing. This mechanism uses auxiliary global contexts to guide local expert selection. Moreover, global-guided dynamic compensation fusion alleviates sample-dependent global--local contribution imbalance while preserving global semantic as the primary anchor and adaptively incorporating expert-refined local cues. Extensive experiments on five multimodal benchmark datasets demonstrate the effectiveness and general applicability of DGC-Net, and ablation studies verify the contribution of each proposed component.

\par\noindent
\textbf{Relevance Statement:}
% This work studies sentiment understanding in user-generated
% Web videos, including video reviews and opinion vlogs.
% Speakers express attitudes through words, vocal tone, and
% facial expressions, whose inconsistencies can make their
% opinions difficult to interpret from transcripts alone.
% By investigating how to integrate these cues, this work
% supports the analysis of sentiment polarity and intensity
% in audiovisual opinions shared on the Web.
% \noindent\textbf{Relevance Statement:}
Online video reviews, opinion vlogs, and video conversations
allow users to express attitudes through spoken words,
vocal tone, and facial expressions.
Conflicting cues across these modalities can make sentiment
difficult to interpret from transcripts alone.
This work investigates cross-modal cue selection and
adaptive global--local fusion to support the analysis of
speakers' sentiment~polarity~and~intensity.
\end{abstract}




\maketitle

% \begin{figure}[t]
%     \centering

%     \begin{subfigure}[t]{0.49\columnwidth}
%         \centering
%         \includegraphics[width=\linewidth]
%         {Figures/motivation_a.png}
%         \caption{Cross-modal cue inconsistency}
%         \label{fig:motivation_a}
%     \end{subfigure}
%     \hfill
%     \begin{subfigure}[t]{0.49\columnwidth}
%         \centering
%         \includegraphics[width=\linewidth]
%         {Figures/motivation_b.png}
%         \caption{Static global--local fusion}
%         \label{fig:motivation_b}
%     \end{subfigure}

%     \caption{(a) Independent intra-modal affective cue selection may
%     produce inconsistent sentiment cues across modalities.
%     (b) Static global--local fusion cannot adapt to the varying
%     reliability of local affective cues, resulting in either
%     insufficient local compensation or excessive local interference.}
%     \label{fig:motivation}
% \end{figure}

\begin{figure}[t]
    \centering
    \captionsetup[subfigure]{
        font+=md,
        labelfont=md,
        textfont=md
    }

    \begin{subfigure}[t]{0.49\columnwidth}
        \centering
        \includegraphics[width=\linewidth]
        {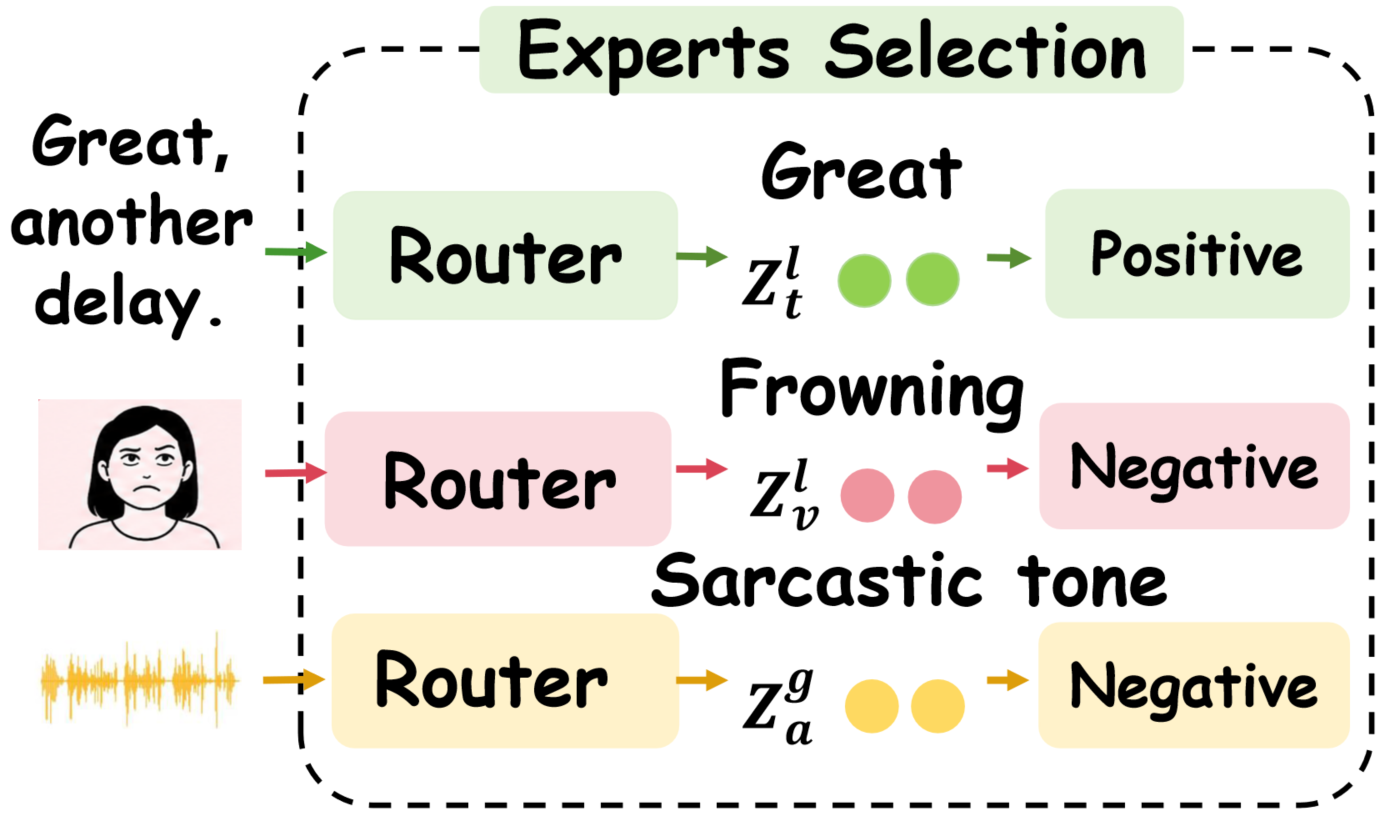}
        \caption{Cross-modal cue inconsistency}
        \label{fig:motivation_a}
    \end{subfigure}
    \hfill
    \begin{subfigure}[t]{0.49\columnwidth}
        \centering
        \includegraphics[width=\linewidth]
        {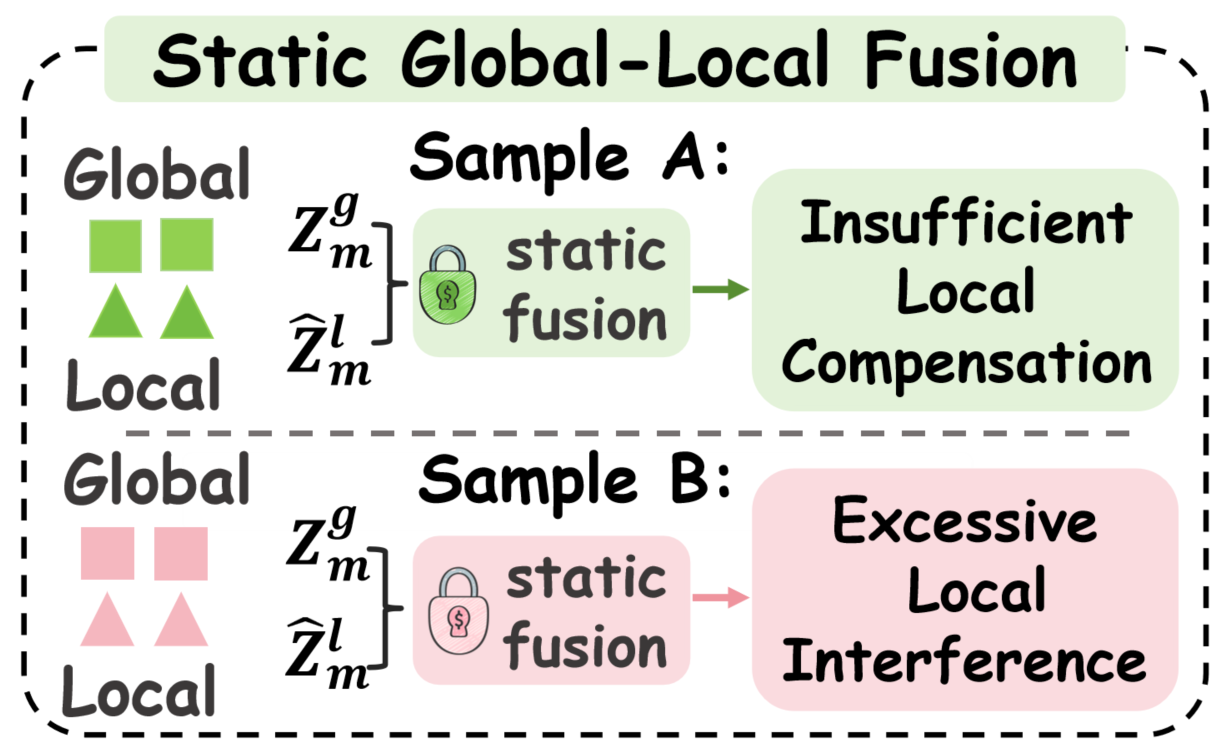}
        \caption{Static global--local fusion}
        \label{fig:motivation_b}
    \end{subfigure}

    \caption{(a) Independent intra-modal affective cue selection may
    produce inconsistent sentiment cues across modalities.
    (b) Static global--local fusion cannot adapt to the varying
    reliability of local affective cues, resulting in either
    insufficient local compensation or excessive local interference.}
    \label{fig:motivation}
\end{figure}

\section{Introduction}

\bl{Online video reviews, personal vlogs, and other forms of
audiovisual communication allow people to express their
opinions~\cite{zadeh2016mosi,zadeh2018mosei}.
Understanding the sentiment conveyed in these videos helps
characterize how users evaluate products, services, and social
topics, supporting the analysis of opinions expressed in online
communities. Such understanding requires identifying both the
polarity and intensity of a speaker's attitude, which are not
always explicit in the transcript. For example, an apparently
positive statement may convey dissatisfaction when accompanied
by a sarcastic tone or a contrasting facial
expression~\cite{castro2019mustard}.
Multimodal sentiment analysis (MSA) therefore integrates textual,
acoustic, and visual information to interpret these
expressions~\cite{tsai2019mult}.
A central challenge is determining how to combine potentially
conflicting cues while retaining the context needed to
understand the speaker's overall sentiment.}

Existing studies mainly address this challenge from two complementary perspectives. The first focuses on selecting and refining informative local affective cues within each modality~\cite{sun2022cubemlp,ma2023prisa}. The second models hierarchical interactions between global contexts and fine-grained local
representations~\cite{lin2023msg,zhuang2024glomo}. The former determines which local cues should be retained, whereas the latter determines how these cues should complement global affective semantics. Despite substantial progress, both directions still face important limitations in cross-modal affective modeling.

Local affective cue modeling methods commonly employ attention, gating, or expert routing to emphasize informative segments and suppress modality-specific noise. 
Recent Mixture-of-Experts (MoE) approaches further introduce specialized experts to refine heterogeneous local cues. However, as depicted in Figure~\ref{fig:motivation}(a), their routing decisions are generally made within each modality independently of the other modalities.
 
Global contexts from other modalities are rarely considered during local expert selection.
Consequently, a modality-specific router may overemphasize noisy cues or select local representations whose affective polarity conflicts with the broader multimodal context.
Thus, local cue refinement should be guided by cross-modal global evidence rather than relying exclusively on the current modality.

Hierarchical fusion methods incorporate global affective contexts to alleviate inconsistencies among modality-specific local representations~\cite{lin2023msg,zhuang2024glomo}.Nevertheless, existing methods determine the relative contributions of global and local representations implicitly through shared projections or fusion parameters. As illustrated in Figure~\ref{fig:motivation}(b), such 
% mechanisms do not 
implicit fusion cannot
explicitly adapt the global-local balance to the sample-dependent reliability of local cues.
Informative local evidence may consequently be underutilized, whereas unreliable local cues may introduce excessive interference.
An effective fusion mechanism should therefore preserve stable global semantics while adaptively incorporating reliable local information as complementary evidence for sentiment prediction.

% \begin{figure}[t]
%     \centering

%     \begin{subfigure}[t]{0.49\columnwidth}
%         \centering
%         \includegraphics[width=\linewidth]
%         {Figures/motivation_a.png}
%         \caption{Cross-modal cue inconsistency}
%         \label{fig:motivation_a}
%     \end{subfigure}
%     \hfill
%     \begin{subfigure}[t]{0.49\columnwidth}
%         \centering
%         \includegraphics[width=\linewidth]
%         {Figures/motivation_b.png}
%         \caption{Static global--local fusion}
%         \label{fig:motivation_b}
%     \end{subfigure}

%     \caption{(a) Independent intra-modal affective cue selection may
%     produce inconsistent sentiment cues across modalities.
%     (b) Static global--local fusion cannot adapt to the varying
%     reliability of local affective cues, resulting in either
%     insufficient local compensation or excessive local interference.}
%     \label{fig:motivation}
% \end{figure}

To address the aforementioned challenges, this paper proposes DGC-Net, 
% a Dynamic Gating and Global-Guided Compensation Network for multimodal sentiment analysis
a Cross-Modal Dynamic Routing with Global-Guided Compensation framework in MSA, aiming to resolve the critical problem of fine-grained alignment of multimodal affective data. 
% First, this paper designs 
\bl{It first introduces} a cross-modal contribution-aware dynamic expert routing mechanism.
% , which dynamically 
\bl{For each modality, this mechanism} estimates the sample-dependent contributions of the current local representation and auxiliary global contexts 
\bl{from the other modalities.}
% to overcome the challenge of representational consistency among multimodal affective cues.
\bl{Their contribution-weighted combination guides sparse expert selection, enabling cross-modal evidence to regulate the refinement of modality-specific local cues.
This design improves the consistency of refined local representations without directly mixing auxiliary modality content into the expert outputs.}
% This mechanism overcomes the limitation of isolated decision-making in traditional expert networks, which rely exclusively on a single modality, by leveraging the global semantics of auxiliary modalities to jointly guide the local feature selection of the current modality. 
% Second, this paper 
\bl{Then, DGC-Net further develops} 
% proposes 
a global-guided dynamic compensation fusion mechanism
\bl{to explicitly regulate the contributions of global and local representations.}
% , which explicitly estimates the sample-dependent global--local contribution ratio within each modality. 
\bl{It estimates a sample-specific global–local contribution ratio for each modality and constrains stable global semantics to serve as the primary representation.}
% This mechanism introduces an explicit proportional constraint model during the feature fusion stage to ensure the dominance of global affective semantics in the fusion computation, while allowing refined local information to serve as dynamic affective compensation.
\bl{Expert-refined local cues are then incorporated as adaptive affective compensation according to their reliability. In this way, DGC-Net avoids both insufficient use of informative local cues and excessive interference from unreliable local evidence during multimodal fusion.}

Our main contributions are summarized as follows:

\begin{itemize}
    \item We propose a cross-modal contribution-aware dynamic expert
    routing mechanism that estimates the sample-dependent
    contributions of auxiliary global contexts to guide expert
    selection for the current modality, thereby improving the
    consistency of local affective representations across the three modalities.

    \item We develop a global-guided dynamic compensation fusion
    mechanism that explicitly models the relative contributions of
    global and local representations. It preserves the dominance of
    stable global semantics while using expert-refined local cues as
    adaptive affective compensation.

    \item Extensive experiments are conducted on five multimodal
    benchmarks, including CMU-MOSI, CMU-MOSEI, UR-FUNNY, MUStARD,
    and CHERMA. The experimental results, ablation studies, and
    further analyses validate the effectiveness of the proposed
    components.
\end{itemize}

\begin{figure*}[!t]
    \centering
    \includegraphics[width=0.98\textwidth]
    {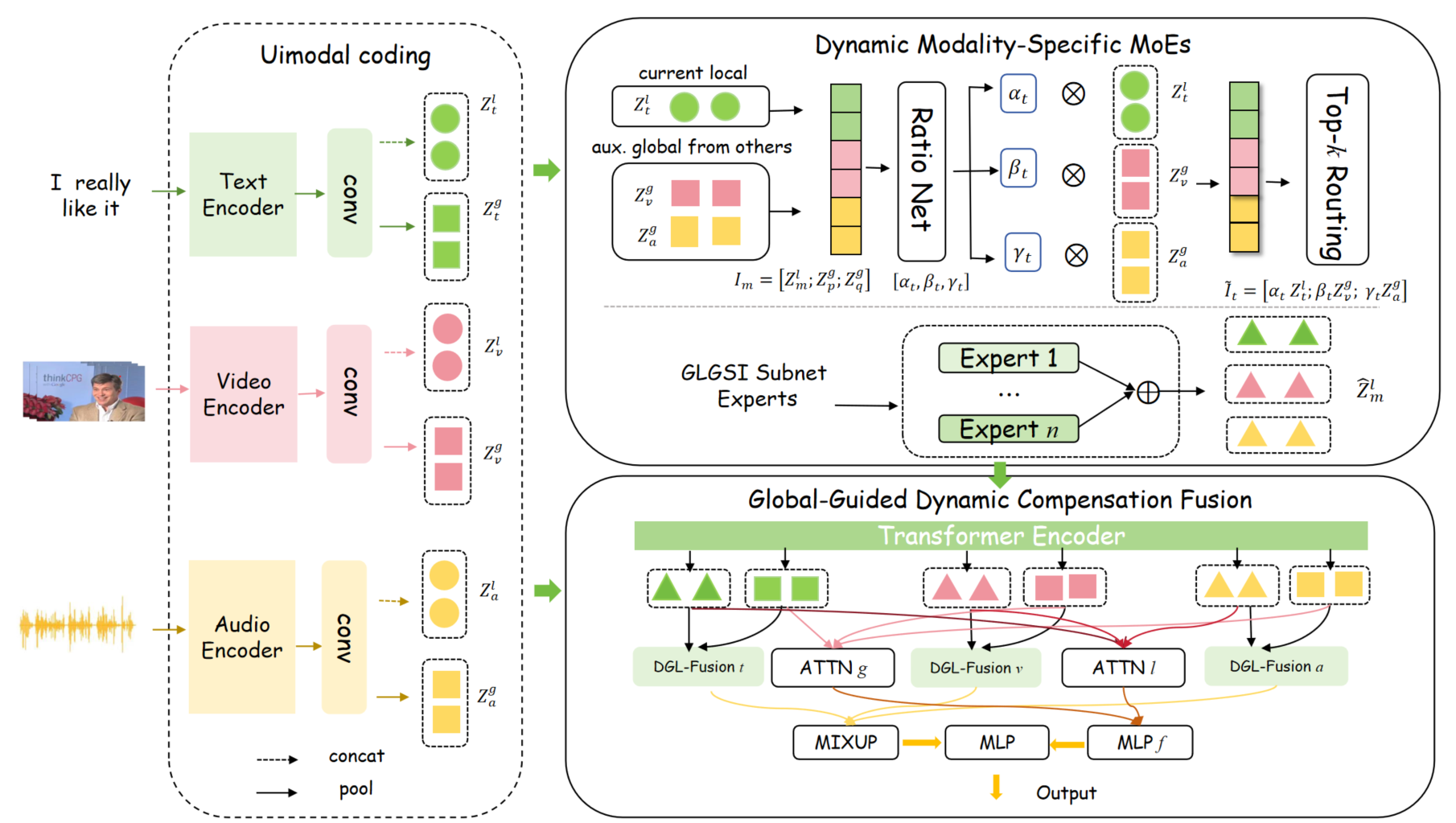}
    \caption{Overall architecture of DGC-Net, consisting of
    unimodal global--local representation, cross-modal
    contribution-aware dynamic expert routing, and global-guided
    dynamic compensation fusion.}
    \label{fig:dgc_framework}
\end{figure*}

\section{Related Work}

Existing fine-grained multimodal affective modeling studies mainly
focus on two directions: dynamic local-cue representation and
hierarchical global--local fusion for multimodal sentiment analysis.

\subsection{Critical Affective Cue Representation and Dynamic Expert Routing}

Multimodal affective cues are often subtle and unevenly distributed.
MulT~\cite{tsai2019mult}, MAG-BERT~\cite{rahman2020magbert}, and
M3SA~\cite{zeng2021m3sa} enhance salient interactions through
attention or gating, while MISA~\cite{hazarika2020misa} and
C-MIB~\cite{mai2022cmib} reduce modality discrepancies and redundant
information. CubeMLP~\cite{sun2022cubemlp},
PriSA~\cite{ma2023prisa}, and MSG~\cite{lin2023msg} further improve
fine-grained affective representations for sentiment prediction.

Sparse MoE~\cite{shazeer2017moe} and Switch
Transformer~\cite{fedus2022switch} establish Top-$k$ expert routing,
while VLMo~\cite{bao2022vlmo} and LIMoE~\cite{mustafa2022limoe}
extend expert specialization to multimodal learning. For affective
modeling, GLoMo~\cite{zhuang2024glomo} introduces modality-specific
experts, MMA~\cite{chen2025mma} combines multimodal adapters with
expert selection, and TEXT~\cite{rao2026text} employs text-routed
sparse experts with temporal alignment. However, existing routers
rarely distinguish the contributions of the current local
representation and auxiliary global contexts. DGC-Net therefore uses
auxiliary global evidence to guide expert selection while keeping the
refined representation modality-specific.

\subsection{Hierarchical Multimodal Fusion and Global--Local Modeling}

Early fusion methods model multimodal interactions through tensor
products, low-rank decomposition, or shared and modality-specific
factors, including TFN~\cite{zadeh2017tfn},
LMF~\cite{liu2018lmf}, and MFM~\cite{tsai2019mfm}. Subsequent
methods improve fine-grained fusion through cross-modal attention,
gating, graph structures, or bidirectional interaction, including
MulT~\cite{tsai2019mult}, MAG-BERT~\cite{rahman2020magbert},
GFN~\cite{mai2020gfn}, M3SA~\cite{zeng2021m3sa}, and
BBFN~\cite{han2021bbfn}.

More recently, CubeMLP~\cite{sun2022cubemlp},
PriSA~\cite{ma2023prisa}, MSG~\cite{lin2023msg}, and
GLoMo~\cite{zhuang2024glomo} jointly model global contexts and local
cues. DPDF-LQ~\cite{zhou2025dpdf} adopts complementary global and
local paths, while EBMC~\cite{he2026ebmc} coordinates modality
contributions according to sample-level reliability. Nevertheless,
within-modality global--local contributions are generally modeled
implicitly, which may underutilize informative local cues or amplify
local noise. DGC-Net explicitly estimates sample-level global--local
ratios, using reliable local cues as compensation while preserving
stable global semantics within each modality.

\section{Methodology}

\subsection{Problem Formulation}

In this section, we formally define the task of multimodal sentiment
analysis. Let $U_m$ denote the input sequence of modality $m$,
where $m\in\mathcal{M}=\{t,v,a\}$ corresponds to the textual,
visual, and acoustic modalities, respectively. The comprehensive
multimodal input is represented as
$\mathcal{U}=\{U_t,U_v,U_a\}$. The primary objective of our work
is to accurately predict sentiment labels or intensities based on these
multimodal sequences. To achieve this, we propose a novel
architecture, DGC-Net, which acts as a mapping function
$f_{\Theta}$ with trainable parameters $\Theta$. Formally, the prediction process is defined for each input as:
\begin{equation}
    \hat{y}=f_{\Theta}(U_t,U_v,U_a),
\end{equation}
where $\hat{y}$ denotes the predicted sentiment output. Specifically,
depending on the downstream dataset, $\hat{y}\in\mathbb{R}$
represents a continuous sentiment intensity score for regression
tasks, or $\hat{y}\in\mathcal{C}$ denotes a predicted category for
classification tasks over a discrete set of classes $\mathcal{C}$ specified by the dataset.

\subsection{Model Overview}

As illustrated in Figure~\ref{fig:dgc_framework}, DGC-Net builds
upon the global--local architecture of
GLoMo~\cite{zhuang2024glomo} and consists of three modules. First,
the Unimodal Coding Module extracts a global representation $Z_m^g$
and a local representation $Z_m^l$ for each modality
$m\in\mathcal{M}={t,v,a}$.

Second, the Dynamic Modality-Specific MoEs Module uses RatioNet to
estimate the contributions of the current local representation and
two auxiliary global representations. Their contribution-weighted
representation guides sparse Top-$k$ gating to select CLGSI Subnet
Experts, producing the refined local representation
$\widehat{Z}_m^l$.

Finally, the Global-Guided Dynamic Compensation Fusion (GDCF)
Module contextualizes the global and refined local representations
through a shared Transformer~\cite{vaswani2017attention}. Its
modality-wise path applies DGL-Fusion followed by MIXUP, while its
feature-wise path summarizes global and local features using
$ATTN_g$, $ATTN_l$, and $MLP_f$. The two outputs are concatenated
for prediction. Overall, the routing and fusion modules respectively
address cross-modal local-cue inconsistency and sample-dependent global--local imbalance within each modality.

\subsection{Unimodal Coding Module}

Expert selection in DGC-Net requires utterance-level context
to interpret fine-grained cues, while subsequent compensation
fusion requires access to both contextual and regional information.
The unimodal coding module therefore constructs global and local
representations for each modality.
Global representations provide auxiliary cross-modal context
for routing and form the global branches in compensation fusion.
Local representations retain regional cues for modality-specific
expert refinement and subsequent local compensation during the fusion stage.

We adopt the encoding scheme of
GLoMo~\cite{zhuang2024glomo} to obtain these representations.
For text, the final-layer \texttt{[CLS]} representation from
BERT is projected into the common hidden space to construct
the global representation.
The local representation is derived from the token features
of the last two BERT layers.
Let $H_t^{-1}$ and $H_t^{-2}$ denote these token representations.
The textual global and local representations are computed as
\begin{equation}
\begin{aligned}
    Z_t^g
    &=
    \operatorname{FF}
    \left(
        \operatorname{CLS}^{-1}(U_t)
    \right),\\
    Z_t^l
    &=
    \operatorname{CONCAT}
    \left(
        \operatorname{AdaMaxPool}_{n}(F_t^l)
    \right),
\end{aligned}
\label{eq:text_representation}
\end{equation}
where $\operatorname{FF}$ denotes a feed-forward projection and
$F_t^l=\operatorname{Conv1D}_t([ H_t^{-1};$
$H_t^{-2}] )$
denotes the convolutionally encoded token features.
The convolution captures local contextual patterns.
Adaptive max pooling then extracts $n$ regional vectors,
which are concatenated to retain regional information
for expert refinement.

For audio and video, a modality-specific convolution projects
the input features, after which a Transformer encoder models
their temporal dependencies.
For $m\in\{a,v\}$, the global and local representations
are computed as follows for the encoded sequence:
\begin{equation}
\begin{aligned}
    Z_m^g
    &=
    \operatorname{MAX}(F_m),\\
    Z_m^l
    &=
    \operatorname{CONCAT}
    \left(
        \operatorname{AdaMaxPool}_{n}(F_m)
    \right),
\end{aligned}
\label{eq:av_representation}
\end{equation}
where
$F_m=\operatorname{TE}_m(\operatorname{Conv1D}_m(U_m))$
denotes the temporally encoded features.
Max pooling aggregates the sequence into an utterance-level
representation, whereas adaptive max pooling retains information
from $n$ temporal regions within each utterance.

For each modality, the coding module produces a global
representation $Z_m^g\in\mathbb{R}^{d}$ and a local
representation $Z_m^l\in\mathbb{R}^{nd}$.
When selecting experts for a target modality $m$,
$Z_m^l$ provides the local evidence to be refined,
while $Z_p^g$ and $Z_q^g$ from the other two modalities
provide auxiliary utterance-level context,
where $\{p,q\}=\mathcal{M}\setminus\{m\}$.
Their relative influence on expert selection should adapt
to each sample.
This motivates the contribution-aware routing mechanism
below, which estimates context weights to guide expert
selection while keeping expert processing specific
to the target modality.

\subsection{Dynamic Modality-Specific MoEs Module}

Conventional modality-specific routers primarily select experts
according to the current local representation. Their decisions may
therefore become unreliable when the current modality is noisy or
affectively ambiguous.To incorporate cross-modal evidence into expert selection, we
propose the Dynamic Modality-Specific MoEs Module. For each modality, RatioNet estimates sample-dependent contribution weights for its local representation and the global representations of the other modalities. These representations
are individually weighted and concatenated to compute expert scores for Top-$k$ selection.
\rev{
Each selected expert, implemented as a CLGSI Subnet~\cite{yang2024clgsi},
independently transforms the target modality's local representation.
The expert outputs
are then aggregated using their corresponding routing
weights according to Eq.~\eqref{eq:expert_combination}
to obtain the refined local representation for the target modality.}

Auxiliary global representations guide the routing decision, while each selected expert processes only the current local representation. This separation introduces cross-modal guidance without changing the modality-specific content. It uses other modalities as semantic evidence for choosing experts, rather than directly injecting their features into the target local representation.

For modality $m$, let $p$ and $q$ denote the other two modalities.
We first concatenate the current local representation and the two
auxiliary global representations as
$I_m=[Z_m^l;Z_p^g;Z_q^g]$. RatioNet then estimates their normalized
contribution coefficients:
\begin{equation}
    [\alpha_m,\beta_m,\gamma_m]
    =
    \operatorname{Softmax}
    \left(
        \operatorname{RatioNet}_m(I_m)
    \right).
\label{eq:contribution_ratio}
\end{equation}

RatioNet is a two-layer MLP with a 64-dimensional hidden layer:
\begin{equation}
    \operatorname{RatioNet}_m(I_m)
    =
    W_{r,2}^{m}
    \operatorname{ReLU}\!\left(
        W_{r,1}^{m}I_m+b_{r,1}^{m}
    \right)
    +b_{r,2}^{m},
\label{eq:ratio_net}
\end{equation}
where $W_{r,1}^{m}\in\mathbb{R}^{64\times d_I}$,
$W_{r,2}^{m}\in\mathbb{R}^{3\times64}$,
and $d_I$ is the dimension of $I_m$.

The Softmax normalization ensures
$\alpha_m+\beta_m+\gamma_m=1$, allowing the three information
sources to be compared on the same scale. Their weighted
concatenation forms the routing representation
\begin{equation}
    \widetilde{I}_m
    =
    [\alpha_m Z_m^l;
     \beta_m Z_p^g;
     \gamma_m Z_q^g].
\label{eq:weighted_routing_input}
\end{equation}

Thus, the auxiliary representations influence expert selection
through their estimated contributions rather than being directly
mixed into the expert content.

Following  Top-$k$ gating~\cite{shazeer2017moe}, the routing weights are obtained as
\begin{equation}
    \boldsymbol{G}_m
    =
    \operatorname{Softmax}
    \left(
        \operatorname{KeepTopK}
        (\boldsymbol{s}_m,k)
    \right),
\label{eq:noisy_topk}
\end{equation}
where
$\boldsymbol{s}_m
=\widetilde{I}_m W_g^m
+\boldsymbol{\epsilon}\odot\delta_m$
denotes the expert scores, and
$\delta_m
=\operatorname{Softplus}(\widetilde{I}_m W_n^m)+\epsilon_0$
is the input-dependent noise scale. Here,
$\boldsymbol{\epsilon}$ is Gaussian noise used only during training,
and $\epsilon_0$ prevents the noise scale from vanishing. The Top-$k$
operation retains only the $k$ highest-scoring experts, encouraging
expert specialization while limiting redundant computation per input sample.

Each selected CLGSI Subnet Expert transforms the vectorized local representation through a bottleneck:
\begin{equation}
    E_i^m(Z_m^l)
    =
    \sigma\!\left(
        W_{2,i}^m
        \operatorname{ReLU}\!\left(
            W_{1,i}^m Z_m^l+b_{1,i}^m
        \right)
        +b_{2,i}^m
    \right).
    \label{eq:subnet_expert}
\end{equation}

The intermediate dimension is set to
$d_b=\max(\lfloor nd/8\rfloor,1)$, forming a compact bottleneck that
suppresses redundant local information before projecting the feature
to dimension $d$. Although all experts share this structure, their
independent parameters allow them to capture different affective
subspaces.

Expert outputs are aggregated using their routing weights:
\begin{equation}
    \widehat{Z}_m^l
    =
    \log
    \left(
        \sum_{i=1}^{N_e}
        G_{m,i}
        \exp
        \left(
            E_i^m(Z_m^l)
        \right)
    \right).
\label{eq:expert_combination}
\end{equation}

Only experts with nonzero Top-$k$ weights contribute to
$\widehat{Z}_m^l$. Because their inputs contain only $Z_m^l$, the
refined representation remains modality-specific, while its
refinement path is selected using cross-modal global evidence from the auxiliary modalities.

\subsection{Global-Guided Dynamic Compensation Fusion Module}

Global representations provide stable utterance-level semantics,
whereas the reliability of fine-grained local cues varies across
samples. Conventional projection-based fusion models their
interaction but does not explicitly control their relative
contributions. We therefore propose the Global-Guided Dynamic
Compensation Fusion Module, which preserves global priority in the
explicit interpolation branch while adaptively incorporating
reliable local evidence.

We arrange the global and expert-refined local representations as
$S=[Z_t^g;\widehat{Z}_t^l;Z_v^g;\widehat{Z}_v^l;
Z_a^g;\widehat{Z}_a^l]$. A shared Transformer contextualizes these global and local representations across modalities:
\begin{equation}
    \{
    \overline{Z}_m^g,
    \overline{Z}_m^l
    \}_{m\in\mathcal{M}}
    =
    \operatorname{Transformer}(S).
\label{eq:shared_transformer}
\end{equation}

This contextualization allows each global--local pair to incorporate
complementary cross-modal information before its contribution ratio
is estimated for dynamic fusion.

For each modality, a shared ratio network predicts the contribution
of its global representation:
\begin{equation}
    \lambda_m
    =
    \frac{1}{2}
    +
    \frac{1}{2}
    \sigma
    \left(
        \operatorname{MLP}_{gl}
        ([\overline{Z}_m^g;\overline{Z}_m^l])
    \right).
\label{eq:global_ratio}
\end{equation}

The resulting coefficient satisfies
$\lambda_m\in(0.5,1)$. It adapts to the sample-dependent reliability
of local cues while ensuring that the global representation receives
no less weight than the local representation in the explicit
interpolation branch.

For modality $m$, dynamic compensation yields
\begin{equation}
    D_m =
    \lambda_m\overline{Z}_m^g
    +(1-\lambda_m)\overline{Z}_m^l
    +\frac{1}{2}\phi_m
    ([\overline{Z}_m^g;\overline{Z}_m^l]).
\label{eq:dgl_fusion}
\end{equation}

The first two terms explicitly control the global--local
contributions, while $\phi_m$ preserves their nonlinear complementary
interaction.
Thus, GDCF separates controlled contribution estimation from
residual nonlinear complementarity instead of relying only on
implicit projection-based fusion.

Global and local attention subsequently summarize the three
modalities at the corresponding representation levels:
\begin{equation}
\begin{aligned}
    (\boldsymbol{w}^g,Z^g)
    &=
    \operatorname{ATTN}_g
    (\overline{Z}_t^g,
     \overline{Z}_v^g,
     \overline{Z}_a^g),\\
    (\boldsymbol{w}^l,Z^l)
    &=
    \operatorname{ATTN}_l
    (\overline{Z}_t^l,
     \overline{Z}_v^l,
     \overline{Z}_a^l).
\end{aligned}
\label{eq:global_local_attention}
\end{equation}

The summaries form the feature-wise fusion path, while global attention weights aggregate the compensated representations:
\begin{equation}
\begin{aligned}
    Z_f
    &=
    \operatorname{MLP}_f([Z^g;Z^l]),\\
    Z_d
    &=
    \operatorname{MIXUP}(D_t,D_v,D_a)
    =
    \sum_{m\in\mathcal{M}}w_m^gD_m.
\end{aligned}
\label{eq:two_fusion_paths}
\end{equation}

The MIXUP operation uses the global attention weights because global
contexts provide a relatively stable basis for estimating modality
importance. Finally, the outputs of the two complementary fusion
paths are combined for sentiment prediction:
\begin{equation}
    \hat{y}
    =
    \operatorname{MLP}([Z_d;Z_f]).
\label{eq:prediction}
\end{equation}

%\wyh{These two paths respectively provide modality-level compensated
Evidence and feature-level global--local summaries, making the final
prediction consistent with both design motivations.

\subsection{Learning Objective}

DGC-Net jointly optimizes task performance, expert utilization, and
contribution estimation. For a batch of $B$ samples, the task loss is
defined according to the prediction target:
\begin{equation}
    \mathcal{L}_{\mathrm{task}}
    =
    \begin{cases}
        \dfrac{1}{B}
        \displaystyle\sum_{b=1}^{B}
        \left|y_b-\hat{y}_b\right|,
        & \text{regression},\\[2mm]
        \operatorname{CE}
        \left(
            \boldsymbol{y},
            \widehat{\boldsymbol{y}}
        \right),
        & \text{classification}.
    \end{cases}
\label{eq:task_loss}
\end{equation}

Accordingly, mean absolute error is used for sentiment regression,
whereas cross-entropy is used for classification.

For modality $m$, let $G_{m,b,i}$ denote the routing weight assigned
to expert $i$ for sample $b$. The importance of expert $i$ is defined
as
$\operatorname{Imp}_{m,i}=\sum_{b=1}^{B}G_{m,b,i}$, while its load is
defined as
$\operatorname{Load}_{m,i}
=\sum_{b=1}^{B}\Pr(i\in\operatorname{TopK}_m(b))$.
The batch-level mean contribution vector is
$\overline{\boldsymbol{r}}_m
=\frac{1}{B}\sum_{b=1}^{B}
[\alpha_{m,b},\beta_{m,b},\gamma_{m,b}]$.
Using these quantities, the routing regularization is defined as
\begin{equation}
    \mathcal{L}_{\mathrm{route}}^m
    =
    \operatorname{CV}(\operatorname{Imp}_m)
    +
    \operatorname{CV}(\operatorname{Load}_m)
    +
    \mu\operatorname{CV}(\overline{\boldsymbol{r}}_m).
\label{eq:routing_regularization}
\end{equation}

The first two terms balance the accumulated routing weights and
expected expert loads, respectively. The third term prevents RatioNet
from consistently assigning most of its contribution to a single
information source, and $\mu$ controls its strength.

For a nonnegative vector $\boldsymbol{x}$, the coefficient of variation is
\begin{equation}
    \operatorname{CV}(\boldsymbol{x})
    =
    \frac{
        \sqrt{\operatorname{Var}(\boldsymbol{x})}
    }{
        \operatorname{Mean}(\boldsymbol{x})+\epsilon
    },
\label{eq:coefficient_variation}
\end{equation}
where $\epsilon$ is a small constant for numerical stability.

The overall training objective combines the task loss with the
routing regularization of all three modalities:
\begin{equation}
    \mathcal{L}
    =
    \mathcal{L}_{\mathrm{task}}
    +
    \lambda_{\mathrm{moe}}
    \sum_{m\in\mathcal{M}}
    \mathcal{L}_{\mathrm{route}}^m,
\label{eq:overall_objective}
\end{equation}
where $\lambda_{\mathrm{moe}}$ controls the overall contribution of
routing regularization.
This learning objective optimizes sentiment prediction
while discouraging routing collapse, which helps keep the dynamic
expert module effective across heterogeneous samples.

\begin{table*}[!t]
\caption{Results on CMU-MOSI and CMU-MOSEI. For Acc-2 and F1,
the values before/after ``/'' correspond to the negative/non-negative
and negative/positive settings, respectively. Bold and underlined
values denote the best and second-best results. $\dagger$ denotes
our reproduced results with checkpoints selected by validation MAE.}
\label{tab:sentiment_results}
\centering
\small
\setlength{\tabcolsep}{3pt}
\begin{tabular*}{\textwidth}{@{\extracolsep{\fill}}l*{10}{c}@{}}
\toprule
& \multicolumn{5}{c}{CMU-MOSI}
& \multicolumn{5}{c}{CMU-MOSEI} \\
\cmidrule(lr){2-6}
\cmidrule(lr){7-11}
Method
& MAE$\downarrow$
& Corr$\uparrow$
& Acc-2$\uparrow$
& F1$\uparrow$
& Acc-7$\uparrow$
& MAE$\downarrow$
& Corr$\uparrow$
& Acc-2$\uparrow$
& F1$\uparrow$
& Acc-7$\uparrow$ \\
\midrule

TFN
& 0.901 & 0.698 & --/80.8 & --/80.7 & 34.9
& 0.593 & 0.700 & --/82.5 & --/82.1 & 50.2 \\

LMF
& 0.917 & 0.695 & --/82.5 & --/82.4 & 33.2
& 0.623 & 0.677 & --/82.0 & --/82.1 & 48.0 \\

MFM
& 0.877 & 0.706 & --/81.7 & --/81.6 & 35.4
& 0.568 & 0.717 & --/84.4 & --/84.3 & 51.3 \\

GFN
& 0.736 & 0.790 & --/84.3 & --/84.3 & 47.0
& 0.611 & 0.774 & --/85.0 & --/85.0 & 51.8 \\

MulT
& 0.767 & \underline{0.799} & --/83.7 & --/83.7 & 41.5
& 0.625 & 0.775 & --/84.7 & --/84.6 & 50.7 \\

MAG-BERT
& 0.790 & 0.769 & --/83.5 & --/83.5 & 42.9
& 0.602 & 0.778 & --/85.0 & --/85.0 & 51.9 \\

M3SA
& 0.730 & 0.793 & --/85.3 & --/85.3 & 45.5
& 0.599 & 0.781 & --/85.2 & --/85.1 & 52.5 \\

ICCN
& 0.860 & 0.710 & --/83.0 & --/83.0 & 39.0
& 0.565 & 0.713 & --/84.2 & --/84.2 & 51.6 \\

CubeMLP
& 0.770 & 0.767 & --/85.6 & --/85.5 & 45.5
& 0.529 & 0.760 & --/85.1 & --/84.5 & 54.9 \\

MISA
& 0.783 & 0.761 & 81.8/83.4 & 81.7/83.6 & 42.3
& 0.555 & 0.756
& \underline{83.6}/85.5
& 83.8/85.3
& 53.2 \\

BBFN
& 0.776 & 0.755 & --/84.3 & --/84.3 & 45.0
& 0.529 & 0.767 & --/86.2 & --/86.1 & 54.8 \\

PriSA
& \underline{0.714} & 0.792
& 83.4/85.5 & 83.2/85.5 & 47.3
& \textbf{0.523} & 0.772
& 82.8/85.9 & 83.2/85.9 & 54.7 \\

C-MIB
& 0.728 & 0.793 & --/85.2 & --/85.2
& \underline{48.2}
& 0.584 & \textbf{0.789}
& --/86.2 & --/86.2 & 53.0 \\

MSG
& 0.748 & 0.782 & --/85.7 & --/85.6 & 45.3
& 0.583 & \underline{0.787}
& --/85.4 & --/85.4 & 52.8 \\

MMA-BERT$^{\dagger}$
& 0.740 & 0.782
& 82.1/83.7 & 82.1/83.8 & 45.3
& 0.537 & 0.764
& 82.5/84.8 & 82.9/84.8 & 53.8 \\

DPDF-LQ$^{\dagger}$
& 0.748 & 0.774
& 80.5/81.9 & 80.5/81.9 & 44.2
& 0.547 & 0.768
& 82.8/85.7 & 83.1/85.6 & 51.5 \\

\midrule

GLoMo
& 0.718 & 0.782
& \underline{84.1}/\textbf{86.7}
& \underline{83.9}/\textbf{86.6}
& \textbf{48.3}
& 0.539 & 0.771
& \textbf{83.7}/\underline{86.5}
& \underline{84.0}/\underline{86.4}
& \underline{55.0} \\

\textbf{DGC-Net}
& \textbf{0.698}
& \textbf{0.800}
& \textbf{84.7}/\underline{86.4}
& \textbf{84.1}/\underline{86.4}
& 45.1
& \underline{0.526}
& 0.782
& 82.3/\textbf{87.0}
& \textbf{84.4}/\textbf{87.0}
& \textbf{55.7} \\

\bottomrule
\end{tabular*}
\end{table*}

\section{Experiments}

\subsection{Experimental Settings}

\paragraph{Datasets.}
We conduct the main experiments on
CMU-MOSI~\cite{zadeh2016mosi} and
CMU-MOSEI~\cite{zadeh2018mosei}, two widely used benchmarks
for multimodal sentiment analysis. Both datasets contain aligned
textual, acoustic, and visual signals with sentiment annotations
ranging from $-3$ to $+3$. CMU-MOSI provides a relatively compact
benchmark, whereas CMU-MOSEI contains substantially more speakers
and utterances and is therefore used to evaluate scalability and
generalization. We follow the standard train, validation, and test
partitions used by GLoMo~\cite{zhuang2024glomo}. Experiments on
UR-FUNNY~\cite{hasan2019urfunny},
MUStARD~\cite{castro2019mustard}, and
CHERMA~\cite{sun2023cherma} are reported in the supplementary
material.

\paragraph{Features and Evaluation Protocol.}
For a controlled comparison, all methods use the same pre-extracted
and word-aligned features as GLoMo. Textual embeddings are obtained
using BERT~\cite{devlin2019bert}, while the acoustic and visual
streams use COVAREP~\cite{degottex2014covarep} and
OpenFace~\cite{baltrusaitis2018openface} features, respectively.
We report MAE and Pearson correlation (Corr) for sentiment
regression, seven-class accuracy (Acc-7) for fine-grained sentiment
prediction, and binary accuracy (Acc-2) and F1 for polarity
recognition. For Acc-2 and F1, we report both the
negative/non-negative and negative/positive settings.

\paragraph{Implementation Details.}
DGC-Net is implemented in PyTorch and optimized using AdamW on an
NVIDIA RTX 3090 GPU. Each modality-specific MoE contains three
CLGSI Subnet Experts and adopts Top-$2$ routing. The routing hidden
dimension, bottleneck ratio, and dropout rate are set to $64$,
$1/8$, and $0.3$, respectively. The regularization coefficients
are set to $\lambda_{\mathrm{moe}}=10^{-2}$ and $\mu=0.1$.
The learning rate and common hidden dimension are adjusted for each
dataset following GLoMo. Model selection is performed using
validation performance without access to the test set. Additional
preprocessing and training details are provided in the supplementary material for reproducing the experiments.

\begin{table}[!t]
\caption{Ablation results on CMU-MOSI and CMU-MOSEI
under the negative/positive setting.
All values are percentages.}
\label{tab:main_ablation}
\centering
\small
\setlength{\tabcolsep}{2pt}
\renewcommand{\arraystretch}{1.0}
\begin{tabularx}{\columnwidth}{
@{}>{\raggedright\arraybackslash}Xcccc@{}
}
\toprule
& \multicolumn{2}{c}{CMU-MOSI}
& \multicolumn{2}{c}{CMU-MOSEI} \\
\cmidrule(lr){2-3}
\cmidrule(lr){4-5}
Variant
& Acc-2$\uparrow$ & F1$\uparrow$
& Acc-2$\uparrow$ & F1$\uparrow$ \\
\midrule

\multicolumn{5}{c}{\textit{Role of GDCF}} \\
w/o DGL-Fusion
& 83.51 & 83.28
& 86.35 & 86.26 \\
Static Equal Ratio
& 85.19 & 85.18
& 85.94 & 85.94 \\
Static Global-heavy Ratio
& 83.05 & 83.12
& 85.91 & 85.87 \\
No Global Weight Constraint
& 83.97 & 83.96
& 85.88 & 85.85 \\

\midrule
\multicolumn{5}{@{}p{\columnwidth}@{}}{
\centering\arraybackslash
\textit{Role of Dynamic Modality-Specific MoEs}} \\
w/o Dynamic MoEs
& 83.05 & 82.88
& 86.35 & 86.29 \\
w/o Cross-modal Guidance
& 83.66 & 83.64
& 85.97 & 85.94 \\
MLP Expert Replacement
& 84.27 & 84.32
& 86.08 & 86.03 \\
Random Routing
& 84.12 & 84.13
& 86.35 & 86.35 \\
Uniform Routing
& 85.80 & 85.73
& 85.28 & 85.32 \\

\midrule
\multicolumn{5}{c}{\textit{Global--Local Branch Controls}} \\
Global Branch Only
& 84.27 & 84.31
& 86.02 & 86.02 \\
Local Branch Only
& 84.27 & 84.25
& 86.22 & 86.18 \\

\midrule
\multicolumn{5}{c}{\textit{Capacity Control}} \\
Parameter-matched MLP
& 81.98 & 82.06
& 86.16 & 86.16 \\

\midrule
\multicolumn{5}{c}{\textit{Unimodal Controls}} \\
Text Only
& 83.97 & 83.95
& 85.41 & 85.43 \\
Audio Only
& 49.62 & 48.45
& 64.75 & 59.13 \\
Vision Only
& 51.76 & 51.51
& 62.40 & 57.36 \\

\midrule
\textbf{DGC-Net (Full)}
& \textbf{86.41} & \textbf{86.38}
& \textbf{86.96} & \textbf{86.95} \\
\bottomrule
\end{tabularx}

\end{table}

\subsection{Overall Performance}

Table~\ref{tab:sentiment_results} reports the overall results on
CMU-MOSI and CMU-MOSEI. DGC-Net achieves the best MAE and Corr on
CMU-MOSI, demonstrating its effectiveness in modeling continuous
sentiment intensity. On the larger CMU-MOSEI benchmark, it achieves
leading Acc-7 and F1 results while maintaining competitive
regression performance. These results show that DGC-Net performs
consistently across sentiment regression, fine-grained
classification, and binary polarity recognition on these sentiment benchmarks.

The gains are not uniform across all metrics. DGC-Net trails GLoMo
on CMU-MOSI Acc-7 and slightly on the negative/positive Acc-2 and
F1 metrics, while its CMU-MOSEI Corr and negative/non-negative
Acc-2 remain below the best results. This suggests that improvements
in continuous estimation do not always transfer directly to
fine-grained discretization or neutral-boundary classification on the evaluated sentiment benchmarks.

Compared with the closely related GLoMo model, DGC-Net provides
clear improvements in regression and most polarity recognition
metrics under the same features and dataset partitions. This
supports the effectiveness of using auxiliary global contexts to
guide local expert selection and dynamically controlling the
contribution of expert-refined local evidence during global--local
fusion. Detailed comparisons with different baseline categories and
metric-specific analyses are provided in the supplementary material.

\begin{figure*}[!t]
    \centering
    \begin{minipage}[t]{0.46\textwidth}
        \centering
        \vspace{0pt}
        \includegraphics[width=\linewidth]
        {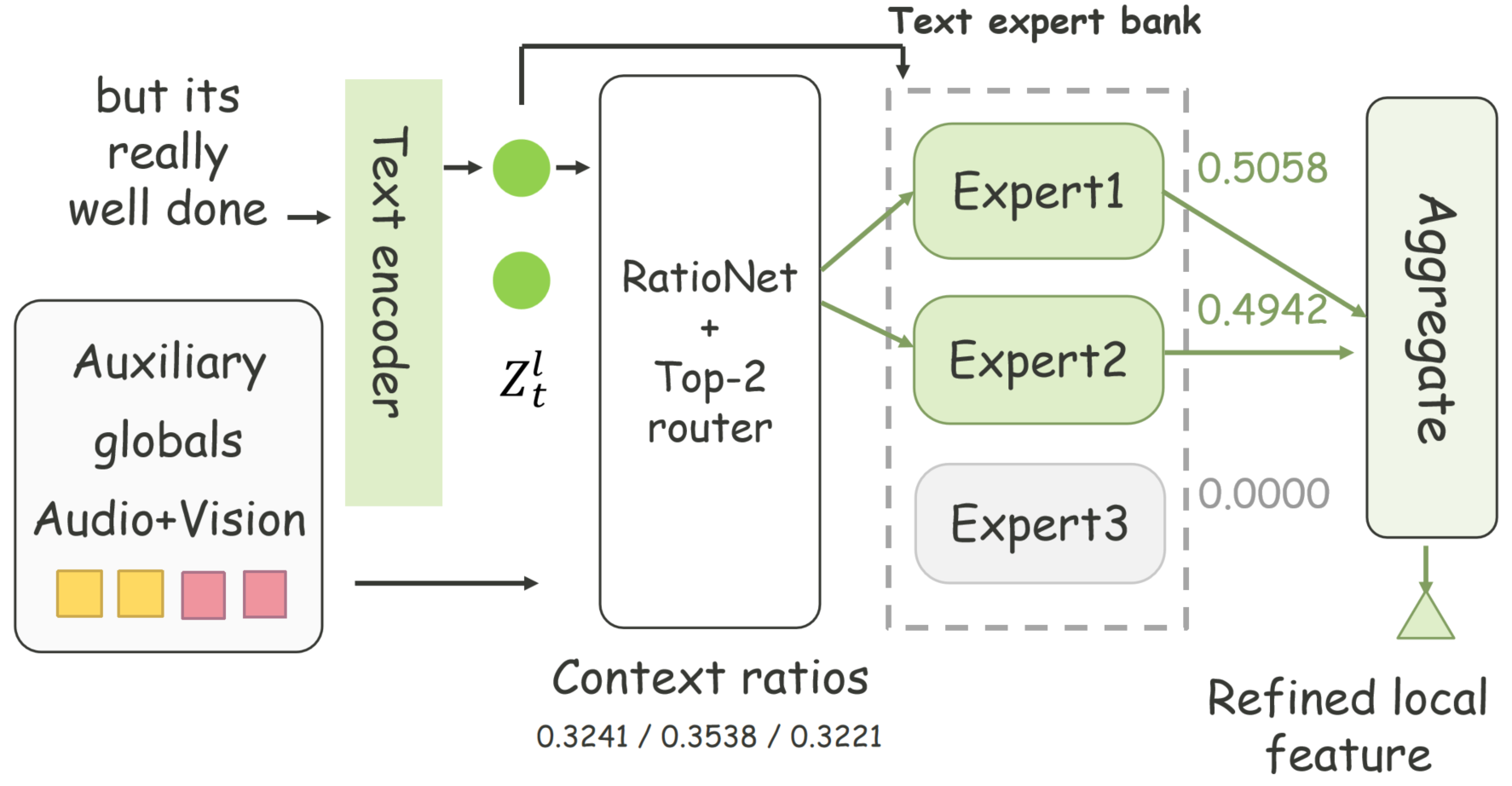}
        \par
        {\small (a) CMU-MOSI sample 130}
    \end{minipage}
    \hfill
    \begin{minipage}[t]{0.46\textwidth}
        \centering
        \vspace{0pt}
        \includegraphics[width=\linewidth]
        {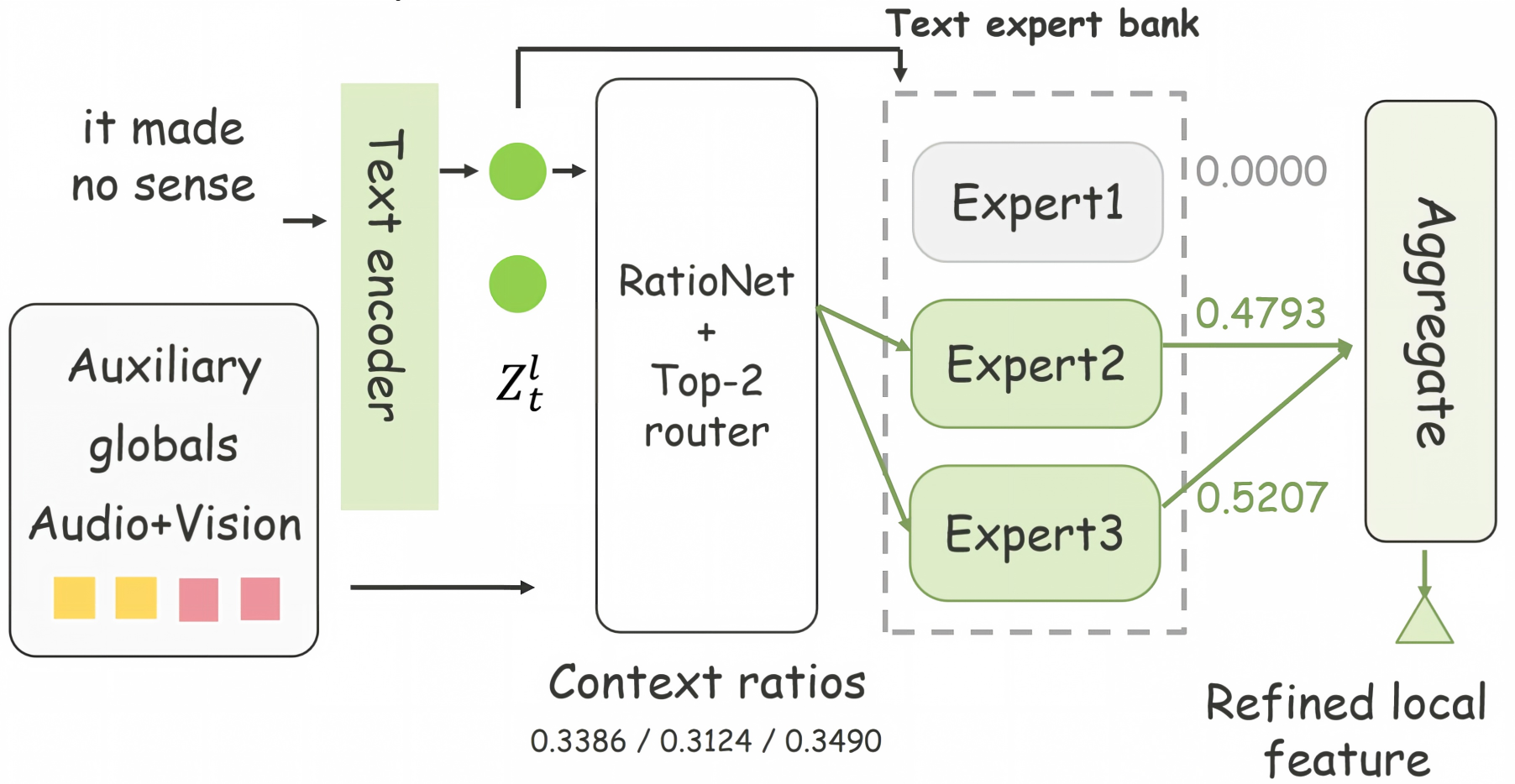}
        \par
        {\small (b) CMU-MOSI sample 125}
    \end{minipage}

    \caption{Sample-dependent text expert routing on CMU-MOSI
    with $n=3$ experts and Top-$k=2$ selection.
    Green and gray boxes indicate active and inactive experts,
    respectively. Numbers beside experts denote gating weights.
    Context ratios follow the local-text, global-audio,
    and global-vision order.}
    \label{fig:routing_cases}
\end{figure*}

\subsection{Ablation Study}

To identify the sources of DGC-Net's performance gains, we conduct
four controlled ablations targeting its fusion, routing, and expert
designs. For GDCF, we replace DGL-Fusion to examine explicit dynamic
global--local compensation. For the Dynamic Modality-Specific MoEs
Module, we remove the complete module and then separately evaluate
cross-modal routing guidance and the CLGSI Subnet Expert
architecture. All variants use the same input features, dataset
partitions, and optimization settings as the full model, with only
the evaluated component changed.

\textit{Role of DGL-Fusion.}
DGL-Fusion is the proposed dynamic compensation component of GDCF.
We replace it with direct MLP fusion while retaining the remaining
operations in GDCF. This replacement reduces Acc-2 and F1 on both
datasets and causes degradation across most metrics, especially on
CMU-MOSI. These results demonstrate that explicitly estimating
sample-dependent global--local contributions is more effective than
learning their relationship implicitly through an MLP.

\textit{Role of Dynamic Modality-Specific MoEs.}
We remove the complete Dynamic Modality-Specific MoEs Module and
feed the original local representations directly into the subsequent
fusion stage. This variant causes the largest Acc-2 and F1 reductions
on CMU-MOSI and also weakens polarity recognition on CMU-MOSEI.
The results demonstrate that modality-specific expert refinement is
important for filtering noisy local cues before global--local fusion.

\textit{Role of Cross-modal Guidance.}
We remove RatioNet and the two auxiliary global representations
while retaining the sparse router, CLGSI Subnet Experts, and GDCF.
Expert selection therefore depends only on the current local
representation. This variant causes substantial degradation across
the CMU-MOSI metrics and the largest Acc-2 and F1 reductions among
the routing variants on CMU-MOSEI. The results verify that auxiliary
global contexts provide useful semantic evidence when local affective
cues are ambiguous or inconsistent across modalities.

\textit{Role of CLGSI Subnet Experts.}
We retain the complete cross-modal contribution-aware routing
mechanism but replace the CLGSI Subnet Experts with conventional
MLP experts. Under identical routing decisions, this replacement
reduces Acc-2 and F1 on both datasets and degrades most of the
remaining metrics. This comparison confirms that the improvements
do not arise from dynamic routing alone and verifies the
effectiveness of the CLGSI Subnet Experts in extracting compact,
task-relevant local representations.

\begin{table}[!t]
\caption{Selected CMU-MOSI cases.
$y$ and $\hat{y}$ denote true and predicted scores.
Error is $|y-\hat{y}|$.
Weights follow text/audio/vision order.}
\label{tab:case_mosi}
\centering
\small
\setlength{\tabcolsep}{0pt}
\renewcommand{\arraystretch}{1.0}

\begin{tabular}{
@{}
p{0.52\columnwidth}
>{\centering\arraybackslash}p{0.16\columnwidth}
>{\centering\arraybackslash}p{0.16\columnwidth}
>{\centering\arraybackslash}p{0.16\columnwidth}
@{}
}
\toprule
Global weights & $y$ & $\hat{y}$ & Error \\
\midrule

\multicolumn{4}{@{}p{\columnwidth}@{}}{
Case 130: but its really well done
} \\
$(0.750,\,0.750,\,0.750)$
& 2.20 & 2.09 & 0.11 \\
\midrule

\multicolumn{4}{@{}p{\columnwidth}@{}}{
Case 125: it made no sense
} \\
$(0.739,\,0.735,\,0.737)$
& -2.00 & -2.04 & 0.04 \\
\midrule

\multicolumn{4}{@{}p{\columnwidth}@{}}{
Case 108: two out of five stars
} \\
$(0.750,\,0.750,\,0.750)$
& -1.60 & 1.49 & 3.09 \\

\bottomrule
\end{tabular}
\end{table}

\begin{figure*}[!t]
    \centering
    \includegraphics[width=0.8\textwidth]
    {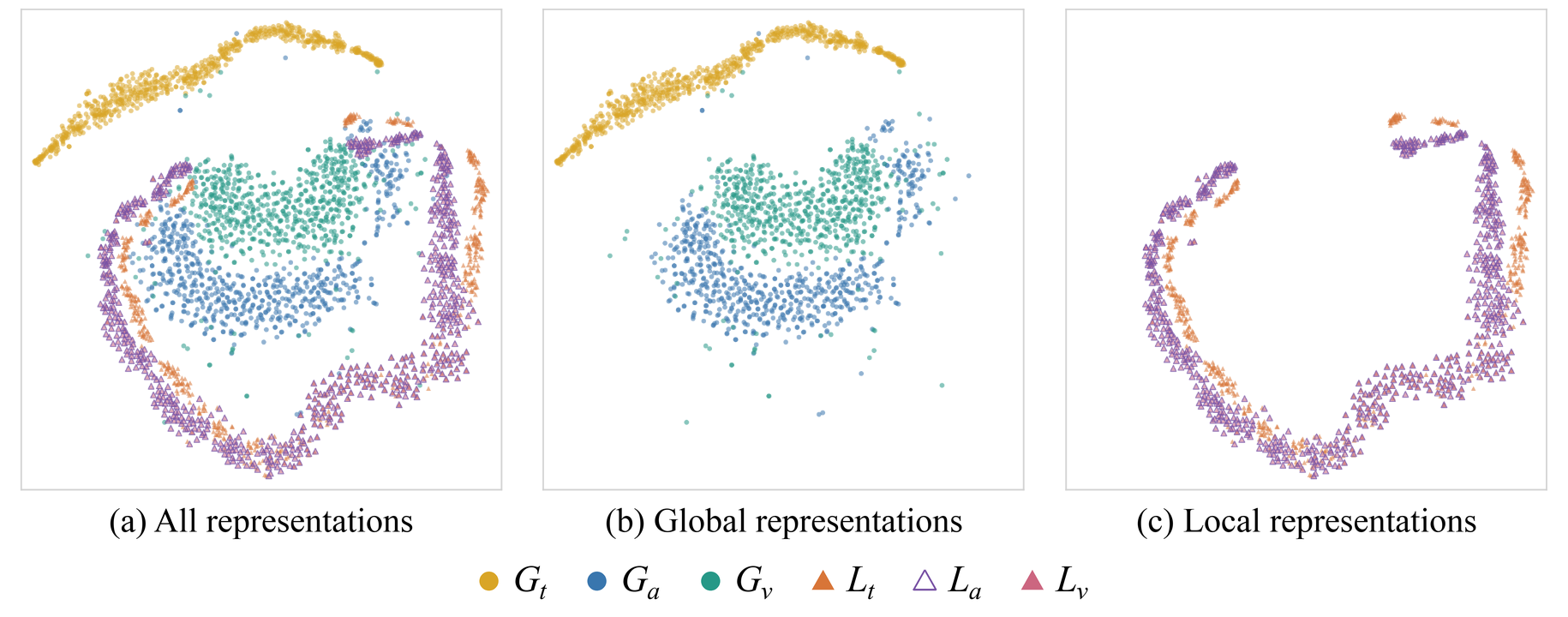}

    \caption{Joint t-SNE visualization on CMU-MOSI.
    Panels (a)--(c) share the same coordinates and display
    all, global, and local representations, respectively.
    $G$ and $L$ denote global and local representations;
    $t$, $a$, and $v$ denote text, audio, and vision.}
    \Description{Three horizontally arranged scatter plots.
    Circular markers represent global features, and triangular
    markers represent local features. Colors distinguish the
    textual, acoustic, and visual modalities.}
    \label{fig:tsne_mosi}
    \end{figure*}

\begin{figure*}[!t]
    \centering
    \includegraphics[width=0.8\textwidth]
    {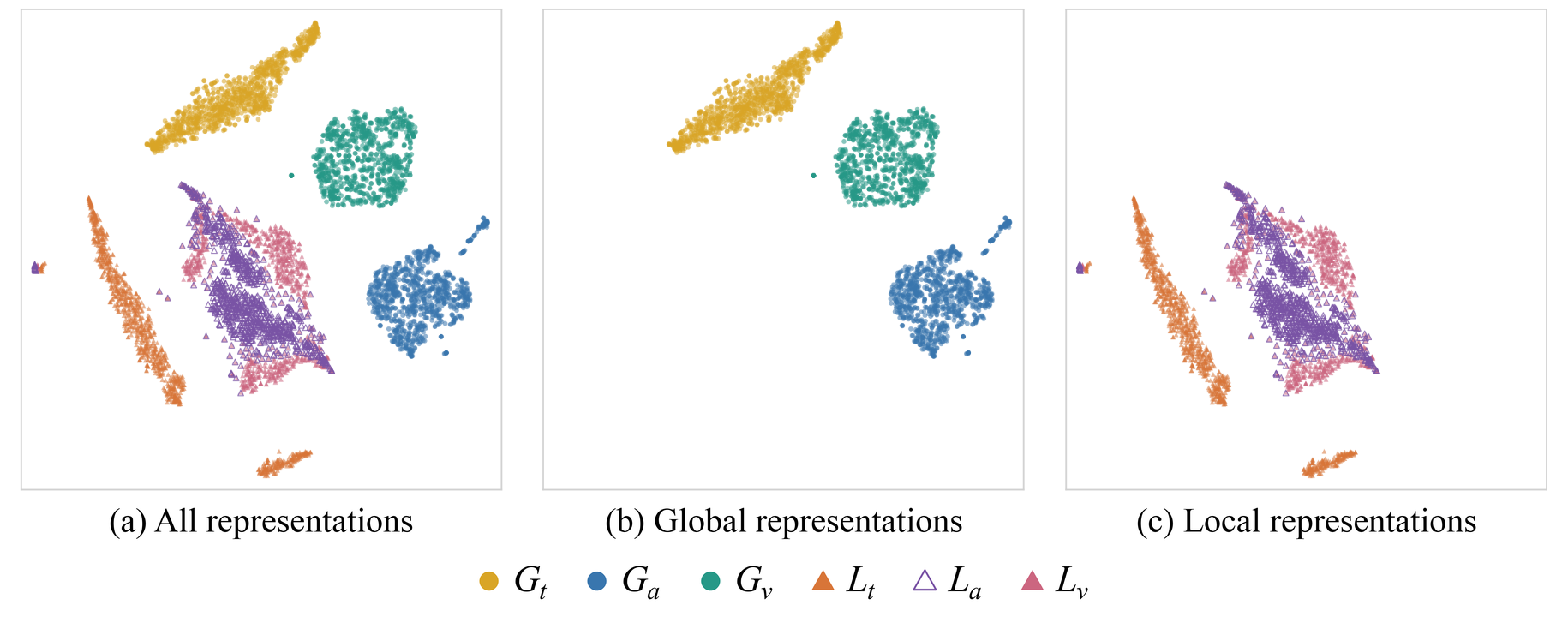}

    \caption{Joint t-SNE visualization on CMU-MOSEI.
    Panels (a)--(c) share coordinates within this dataset
    and display all, global, and local representations,
    respectively. Notation follows
    Figure~\ref{fig:tsne_mosi}.}
    \Description{Three horizontally arranged scatter plots.
    The first contains all six representation groups.
    The second displays global features, and the third
    displays local features at their original projected positions.}
    \label{fig:tsne_mosei}
    \end{figure*}

\subsection{Case Study}

\paragraph{Cases on CMU-MOSI}
Table~\ref{tab:case_mosi} includes two accurate predictions
and one polarity error. The model correctly interprets
``but its really well done'' and ``it made no sense,''
with absolute errors of 0.11 and 0.04, respectively.
However, ``two out of five stars'' receives a positive
prediction despite its negative ground-truth score.
The positive example and the failed example have similar
global weights at the displayed precision, illustrating
that these coefficients alone do not explain prediction
correctness. Due to space limitations, the case study table and corresponding examples for CMU-MOSEI are provided
in the appendix.

% \paragraph{Cases on CMU-MOSEI}
% Table~\ref{tab:case_mosei} shows accurate predictions for
% explicit positive and negative expressions.
% Both examples assign higher global weights to the text
% branch than to the audio and vision branches.
% In contrast, Case 2567 receives a near-zero positive
% prediction despite its negative label, highlighting a
% remaining difficulty in interpreting the exported
% utterance containing both ``enjoy'' and ``did not.''
% These examples illustrate gating behavior and prediction
% limitations without establishing modality reliability.

% \begin{table}[t]
% \caption{Selected CMU-MOSEI cases.
% Notation and weight order follow Table~\ref{tab:case_mosi}.}
% \label{tab:case_mosei}
% \centering
% \normalsize
% \setlength{\tabcolsep}{0pt}
% \renewcommand{\arraystretch}{1.1}

% \begin{tabular}{
% @{}
% p{0.52\columnwidth}
% >{\centering\arraybackslash}p{0.16\columnwidth}
% >{\centering\arraybackslash}p{0.16\columnwidth}
% >{\centering\arraybackslash}p{0.16\columnwidth}
% @{}
% }
% \toprule
% Global weights & $y$ & $\hat{y}$ & Error \\
% \midrule

% \multicolumn{4}{@{}p{\columnwidth}@{}}{
% Case 3918: but it a great color I love
% } \\
% $(0.819,\,0.737,\,0.729)$
% & 2.00 & 2.15 & 0.15 \\
% \midrule

% \multicolumn{4}{@{}p{\columnwidth}@{}}{
% Case 2771: it just so stupid.
% } \\
% $(0.827,\,0.724,\,0.732)$
% & -3.00 & -2.82 & 0.18 \\
% \midrule

% \multicolumn{4}{@{}p{\columnwidth}@{}}{
% Case 2567: enjoy this one I did not it has the same
% } \\
% $(0.752,\,0.736,\,0.729)$
% & -1.33 & 0.05 & 1.38 \\

% \bottomrule
% \end{tabular}
% \end{table}

\paragraph{Sample-dependent expert routing.}
Figure~\ref{fig:routing_cases} further illustrates the routing
behavior of the two correctly predicted MOSI samples.
The text expert bank contains $n$ experts, with $n=3$
and Top-$k=2$ in these experiments.
Sample 130 activates Experts 1 and 2 with weights of
0.5058 and 0.4942, whereas Sample 125 activates Experts 2
and 3 with weights of 0.4793 and 0.5207.
The router combines local text features with auxiliary
audio and visual global contexts, while the selected
experts transform only local text features.
These cases demonstrate sample-dependent expert selection,
but do not establish polarity-specific expert specialization
or a causal link between routing choices and prediction accuracy.

\subsection{Advanced Analysis}

Additional results are provided in the supplementary material,
including experiments on UR-FUNNY, MUStARD, and CHERMA, analyses
of the learned routing and compensation coefficients, and
sensitivity studies on the number of experts and activated
experts. These experiments further evaluate cross-task
generalization, sample-dependent behavior, and routing robustness.
\subsection{Representation Analysis}

Figures~\ref{fig:tsne_mosi} and~\ref{fig:tsne_mosei}
visualize the global and local representations on
CMU-MOSI and CMU-MOSEI, respectively.
For each dataset, the six representation groups are
jointly projected into a two-dimensional space using t-SNE.
The three panels share the same coordinates and display
all representations, global representations, and local
representations, respectively.
The latter two panels filter the joint projection for
visual inspection rather than represent model ablations of individual model components.

On CMU-MOSI, global text representations occupy a region
largely separate from global audio and visual representations,
which partially overlap in this projection.
% On CMU-MOSEI, the three global representation groups appear
% more distinctly separated.
% Local audio and visual representations exhibit substantial
% overlap in both datasets.
% These observations characterize the projected distributions
% at different representation granularities; separation or
% overlap alone does not establish semantic complementarity,
% modality reliability, or improved prediction performance.

\section{Conclusion}

In this paper, we propose DGC-Net to address two critical limitations of global--local modeling in multimodal sentiment analysis: the inconsistency arising from the independent selection of local affective cues, and the sample-dependent reliability of global and local representations. Specifically, DGC-Net initially employs a Unimodal Coding Module to extract both stable global semantics and fine-grained local cues from each modality. Subsequently, it introduces a Dynamic Modality-Specific MoEs Module, where auxiliary global contexts and learned contribution ratios explicitly guide expert routing, empowering the CLGSI Subnet Experts to precisely refine the local representation of the target modality. Finally, a Global-Guided Dynamic Compensation Fusion Module adaptively estimates the relative contributions of global and local information, utilizing stable global semantics as the foundational anchor and reliable local cues as sample-dependent compensation. Extensive experiments on multiple multimodal affective computing benchmarks demonstrate the effectiveness and competitiveness of DGC-Net, and rigorous ablation studies validate the specific contributions of our proposed routing, expert, and dynamic compensation mechanisms for multimodal sentiment prediction. Together, the modules connect cross-modal guidance to local cue refinement and adaptive global--local feature compensation.

In future work, we aim to explore more robust modeling techniques for neutral and fine-grained sentiment boundaries, extend the framework to larger and more heterogeneous multimodal datasets, and optimize the computational overhead of dynamic expert routing for real-world applications.We will further examine how expert selection and
global--local compensation respond to noisy or missing
modalities. Evaluations across video sources and recording
conditions will help clarify the framework's applicability
to sentiment understanding in user-generated Web videos.

% % References
% % Finish all body floats before starting the references.
% \clearpage
% \bibliographystyle{ACM-Reference-Format}
% \bibliography{aaai2027}

% \end{document}

% 正文结束后，统一输出全文参考文献
\clearpage
\bibliographystyle{ACM-Reference-Format}
\bibliography{aaai2027}

% 接着输出附录
\clearpage
\appendix
\input{appendix}

\end{document}

%% file: appendix.tex
% Supplementary content for the combined twelve-page WWW manuscript.
% Keep main.tex as the Overleaf Main document and use \input{appendix}.
% The main file supplies \appendix, packages, and the single bibliography.
% Existing experimental data and both algorithm procedures are retained.

\section{Supplementary Experimental Settings}

\subsection{Datasets and Evaluation}

We use GLoMo's features and partitions~\cite{zhuang2024glomo}.
CMU-MOSI~\cite{zadeh2016mosi} contains $2{,}195$ utterances from
$93$ opinion videos and $89$ speakers, with $1{,}281/229/685$
training/validation/test samples. CMU-MOSEI~\cite{zadeh2018mosei}
contains $3{,}228$ videos from approximately $1{,}000$ speakers,
with $16{,}265/1{,}869/4{,}643$ samples. Both use sentiment labels
in $[-3,3]$. The cleaned UR-FUNNY~\cite{hasan2019urfunny} split
contains $7{,}614/980/994$ utterances; the MUStARD
split~\cite{castro2019mustard} contains $539/68/68$ utterances.
CHERMA~\cite{sun2023cherma} contains $28{,}717$ Chinese utterances
covering seven emotions and uses a $6{:}2{:}2$ split.
Table~\ref{tab:appendix_datasets} summarizes the tasks.

For sentiment regression, we report MAE and Pearson correlation;
Acc-7 evaluates seven-class predictions. Binary Acc-2 and F1 either
retain neutral samples as non-negative or exclude them for the
negative/positive setting. Humor and sarcasm use binary accuracy;
CHERMA uses accuracy and Overall-F1. Hyperparameters and checkpoints
are selected on validation data, with the test set reserved for
final evaluation.

\begin{table}[!ht]
\caption{Tasks and principal evaluation metrics of the five
multimodal benchmarks.}
\label{tab:appendix_datasets}
\centering
\small
\setlength{\tabcolsep}{3pt}
\begin{tabular}{@{}llll@{}}
\toprule
Dataset & Task & Label space & Main metric \\
\midrule
CMU-MOSI & Sentiment & $[-3,3]$ & MAE/Corr \\
CMU-MOSEI & Sentiment & $[-3,3]$ & MAE/Corr \\
UR-FUNNY & Humor & Binary & Acc-2 \\
MUStARD & Sarcasm & Binary & Acc-2 \\
CHERMA & Emotion & Seven classes & Acc/Overall-F1 \\
\bottomrule
\end{tabular}

\end{table}

\subsection{Baseline Models}

For CMU-MOSI and CMU-MOSEI, we compare DGC-Net with four categories
of representative baselines. Tensor-based and factorized fusion
methods include TFN~\cite{zadeh2017tfn},
LMF~\cite{liu2018lmf}, and MFM~\cite{tsai2019mfm}, which model
multimodal interactions through tensor products, low-rank
decomposition, or factorized representations. Cross-modal
interaction methods include GFN~\cite{mai2020gfn},
MulT~\cite{tsai2019mult}, MAG-BERT~\cite{rahman2020magbert},
M3SA~\cite{zeng2021m3sa}, ICCN~\cite{sun2020iccn}, and
BBFN~\cite{han2021bbfn}, which enhance information exchange through
translation, attention, gating, or bidirectional interaction.
Representation learning and information-constraint methods include
CubeMLP~\cite{sun2022cubemlp}, MISA~\cite{hazarika2020misa}, and
C-MIB~\cite{mai2022cmib}. Adaptive and hierarchical fusion methods
include PriSA~\cite{ma2023prisa}, MSG~\cite{lin2023msg}, and
GLoMo~\cite{zhuang2024glomo}, which coordinate modality or
global--local representations during fusion.

We additionally include two recent baselines reported in the main
comparison table: MMA-BERT~\cite{chen2025mma} and
DPDF-LQ~\cite{zhou2025dpdf}. MMA-BERT combines multimodal adapter
modeling with expert selection, providing a comparison with recent
expert-based multimodal representation learning. DPDF-LQ adopts
complementary global and local paths with dynamic fusion, providing
a comparison with recent global--local modeling. For these two
methods, the main table reports our reproduced results, with
checkpoints selected by validation MAE, as indicated by
$\dagger$.

For UR-FUNNY and MUStARD, we follow the evaluation protocol of
GLoMo~\cite{zhuang2024glomo} and include
MISA~\cite{hazarika2020misa} and
MAG-BERT~\cite{rahman2020magbert} with different textual backbones.
We also report AGM and MIL results provided under the same protocol
in the GLoMo appendix~\cite{zhuang2024glomo}.
GLoMo serves as the principal controlled baseline because DGC-Net
uses the same features, dataset partitions, and unimodal coding
pipeline.

For CHERMA, we follow its original benchmark
setting~\cite{sun2023cherma}, which includes Transformer-based early
and late fusion methods and approaches using modality-wise emotion
labels. Because DGC-Net does not require additional unimodal
supervision, its comparison with GLoMo~\cite{zhuang2024glomo}
directly evaluates the effects of cross-modal contribution-aware
dynamic expert routing and Global-Guided Dynamic Compensation
Fusion (GDCF).

\subsection{Implementation}

For the four English benchmarks, inputs use
BERT~\cite{devlin2019bert}, COVAREP~\cite{degottex2014covarep},
and OpenFace~\cite{baltrusaitis2018openface}; acoustic and visual
features are aligned to text. UR-FUNNY and MUStARD concatenate
context with the target utterance. CHERMA uses Chinese BERT-base,
wav2vec, and ResNet-18. Sequences are truncated or zero-padded.
The model uses PyTorch, AdamW, and an RTX 3090.
Transformer layers~\cite{vaswani2017attention} encode context;
RatioNet has $64$ hidden units. CLGSI Subnet
Experts~\cite{yang2024clgsi} use a two-layer transformation with
a $1/8$ bottleneck and Sigmoid output. Dropout is $0.3$ and
$\mu=0.1$. Table~\ref{tab:appendix_implementation} gives the
remaining settings.

\begin{table}[!ht]
\caption{Dataset-specific settings. $L$, $T$, and $d$ denote
encoder layers, maximum sequence length, and hidden dimension.
All datasets use $n=3$ local representations, three experts per
modality, Top-$k=2$, and $\lambda_{\mathrm{moe}}=10^{-2}$.}
\label{tab:appendix_implementation}
\centering
\small
\setlength{\tabcolsep}{3pt}
\begin{tabular}{@{}lccccc@{}}
\toprule
Dataset & Batch & $L$ & $T$ & $d$ & Learning rate \\
\midrule
CMU-MOSI & 240 & 4 & 60 & 48 & $4\times10^{-5}$ \\
CMU-MOSEI & 64 & 3 & 80 & 192 & $1\times10^{-5}$ \\
UR-FUNNY & 220 & 3 & 80 & 112 & $2\times10^{-5}$ \\
MUStARD & 64 & 3 & 70 & 160 & $2\times10^{-5}$ \\
CHERMA & 400 & 7 & 50 & 256 & $2\times10^{-5}$ \\
\bottomrule
\end{tabular}
\end{table}

\section{Algorithms}

Algorithms~\ref{alg:dynamic_moe} and~\ref{alg:dgc_training}
summarize routing and full-model optimization. Auxiliary globals
guide selection, while each activated expert processes only the
target local representation. In GDCF, the bound on $\lambda_m$
favors global information within the interpolation term; it does
not guarantee global dominance of the final prediction.

\begin{algorithm}[!ht]
\caption{Cross-Modal Contribution-Aware Dynamic Expert Routing}
\label{alg:dynamic_moe}
\footnotesize
\begin{algorithmic}[1]
\Require Current local representation $Z_m^l$; auxiliary global
representations $Z_p^g,Z_q^g$; experts
$\{E_i^m\}_{i=1}^{N_e}$; activated-expert number $k$; training
indicator $\tau$
\Ensure Refined local representation $\widehat{Z}_m^l$, routing
weights $\boldsymbol{G}_m$, and contribution ratios
$[\alpha_m,\beta_m,\gamma_m]$

\State $I_m \gets [Z_m^l;Z_p^g;Z_q^g]$

\State $[\alpha_m,\beta_m,\gamma_m] \gets
\operatorname{Softmax}(\operatorname{RatioNet}_m(I_m))$

\State $\boldsymbol{r}_m\gets
[\alpha_m,\beta_m,\gamma_m]$

\State $\widetilde{I}_m \gets
[\alpha_m Z_m^l;\beta_m Z_p^g;\gamma_m Z_q^g]$

\State $\boldsymbol{s}_m \gets \widetilde{I}_m W_g^m$

\If{$\tau$}
    \State $\boldsymbol{\epsilon}\sim
    \mathcal{N}(\boldsymbol{0},\boldsymbol{I})$
    \State $\boldsymbol{s}_m \gets
    \boldsymbol{s}_m+
    \boldsymbol{\epsilon}\odot
    \left[
    \operatorname{Softplus}
    (\widetilde{I}_m W_n^m)+\epsilon_0
    \right]$
\EndIf

\State $\boldsymbol{G}_m \gets
\operatorname{Softmax}\left(
\operatorname{KeepTopK}(\boldsymbol{s}_m,k)
\right)$

\State $\mathcal{S}_m \gets
\{i\mid G_{m,i}>0\}$

\ForAll{$i\in\mathcal{S}_m$}
    \State $E_i^m(Z_m^l)\gets
    \sigma\!\left(
    W_{2,i}^m\operatorname{ReLU}
    (W_{1,i}^m Z_m^l+b_{1,i}^m)+b_{2,i}^m
    \right)$
\EndFor

\State $\widehat{Z}_m^l \gets
\log\!\left(
\sum_{i\in\mathcal{S}_m}
G_{m,i}\exp(E_i^m(Z_m^l))
\right)$

\State \Return $\widehat{Z}_m^l,\boldsymbol{G}_m,
\boldsymbol{r}_m$
\end{algorithmic}
\end{algorithm}

\begin{algorithm}[!ht]
\caption{Forward Propagation and Optimization of DGC-Net}
\label{alg:dgc_training}
\footnotesize
\begin{algorithmic}[1]
\Require Multimodal utterance
$\mathcal{U}=\{U_t,U_v,U_a\}$; training indicator $\tau$; label
$y$ when $\tau$ is true
\Ensure Prediction $\hat{y}$

\ForAll{$m\in\mathcal{M}=\{t,v,a\}$}
    \State $(Z_m^g,Z_m^l)\gets
    \operatorname{UnimodalCoding}_m(U_m)$
\EndFor

\ForAll{$m\in\mathcal{M}$}
    \State Let $\{p,q\}\gets
    \mathcal{M}\setminus\{m\}$

    \State $(\widehat{Z}_m^l,\boldsymbol{G}_m,
    \boldsymbol{r}_m)\gets
    \operatorname{DynamicMoE}_m
    (Z_m^l,Z_p^g,Z_q^g,k,\tau)$
\EndFor

\State $S\gets
[Z_t^g;\widehat{Z}_t^l;
Z_v^g;\widehat{Z}_v^l;
Z_a^g;\widehat{Z}_a^l]$

\State $\{\overline{Z}_m^g,\overline{Z}_m^l\}_{m\in\mathcal{M}}
\gets \operatorname{Transformer}(S)$

\ForAll{$m\in\mathcal{M}$}
    \State $\lambda_m\gets
    \frac{1}{2}+\frac{1}{2}
    \sigma\!\left(
    \operatorname{MLP}_{gl}
    ([\overline{Z}_m^g;\overline{Z}_m^l])
    \right)$

    \State $D_m\gets
    \lambda_m\overline{Z}_m^g+
    (1-\lambda_m)\overline{Z}_m^l+
    \frac{1}{2}\phi_m
    ([\overline{Z}_m^g;\overline{Z}_m^l])$
\EndFor

\State $(\boldsymbol{w}^g,Z^g)\gets
\operatorname{ATTN}_g
(\overline{Z}_t^g,
 \overline{Z}_v^g,
 \overline{Z}_a^g)$

\State $(\boldsymbol{w}^l,Z^l)\gets
\operatorname{ATTN}_l
(\overline{Z}_t^l,
 \overline{Z}_v^l,
 \overline{Z}_a^l)$

\State $Z_f\gets
\operatorname{MLP}_f([Z^g;Z^l])$

\State $Z_d\gets
\operatorname{MIXUP}(D_t,D_v,D_a)
=\sum_{m\in\mathcal{M}}w_m^gD_m$

\State $\hat{y}\gets
\operatorname{MLP}([Z_d;Z_f])$

\If{$\tau$}
    \State $\mathcal{L}\gets
    \mathcal{L}_{\mathrm{task}}(y,\hat{y})+
    \lambda_{\mathrm{moe}}
    \sum_{m\in\mathcal{M}}
    \mathcal{L}_{\mathrm{route}}^m$

    \State Update $\Theta$ by AdamW using
    $\nabla_{\Theta}\mathcal{L}$
\EndIf

\State \Return $\hat{y}$
\end{algorithmic}
\end{algorithm}

\section{Results on Additional Benchmarks}

Table~\ref{tab:humor_sarcasm_results} reports binary accuracy on
UR-FUNNY and MUStARD, using baselines from the GLoMo appendix.
DGC-Net improves GLoMo by $0.79$ and $0.96$ percentage points,
respectively. These aggregate results extend the evaluation to
humor and sarcasm but do not independently verify routing behavior.

Table~\ref{tab:cherma_results} reports CHERMA Overall-F1, the metric
available for earlier baselines. DGC-Net ranks second at
$72.00\%$, exceeding LFMIM by $1.46$ percentage points but
remaining below GLoMo. Its accuracy is $72.16\%$, compared with
GLoMo's $73.74\%$.

\begin{table}[!ht]
\caption{Acc-2 results on UR-FUNNY and MUStARD. Baseline values
are taken from the GLoMo appendix. ``--'' indicates that the
corresponding result was not reported. Bold and underlined values
indicate the best and second-best results, respectively.}
\label{tab:humor_sarcasm_results}
\centering
\small
\setlength{\tabcolsep}{5pt}
\begin{tabular}{@{}lcc@{}}
\toprule
Method & UR-FUNNY$\uparrow$ & MUStARD$\uparrow$ \\
\midrule
MISA (BERT)
& 69.62 & 66.18 \\
MISA (ALBERT)
& 69.82 & 66.18 \\
MAG-BERT (ALBERT)
& 67.20 & 69.12 \\
MAG-BERT (XLNet)
& 72.43 & 76.47 \\
AGM (BERT)
& 65.97 & -- \\
MIL (BERT)
& -- & 76.36 \\
\midrule
GLoMo (BERT)
& \underline{74.95} & \underline{83.86} \\
\textbf{DGC-Net (BERT)}
& \textbf{75.74} & \textbf{84.82} \\
\bottomrule
\end{tabular}

\end{table}

\begin{table}[!ht]
\caption{Overall-F1 results on CHERMA. Baseline values are taken
from the GLoMo appendix. All values are percentages. Bold and
underlined values indicate the best and second-best results,
respectively.}
\label{tab:cherma_results}
\centering
\small
\setlength{\tabcolsep}{12pt}
\begin{tabular}{@{}lc@{}}
\toprule
Method & Overall-F1$\uparrow$ \\
\midrule
TFN   & 68.37 \\
LMF   & 68.23 \\
EFT   & 68.72 \\
LFT   & 69.05 \\
MulT  & 69.24 \\
PMR   & 69.53 \\
LFMIM & 70.54 \\
\midrule
\textbf{GLoMo} & \textbf{73.62} \\
\underline{DGC-Net} & \underline{72.00} \\
\bottomrule
\end{tabular}

\end{table}

\section{Extended Ablation Studies}

\begin{table}[!t]
\caption{Extended ablation results on UR-FUNNY, MUStARD, and
CHERMA. Acc-2 is reported for UR-FUNNY and MUStARD, and overall
accuracy for CHERMA. All values are percentages. Results use
seeds 87, 102, and 20, respectively. Bold values indicate the
highest accuracy in each column, including ties.}
\label{tab:additional_ablation}
\centering
\small
\setlength{\tabcolsep}{2pt}
\renewcommand{\arraystretch}{1.1}
\begin{tabularx}{\columnwidth}{
@{}>{\raggedright\arraybackslash}Xccc@{}
}
\toprule
Variant
& \shortstack{UR-FUNNY\\Acc-2$\uparrow$}
& \shortstack{MUStARD\\Acc-2$\uparrow$}
& \shortstack{CHERMA\\Acc$\uparrow$} \\
\midrule
\multicolumn{4}{@{}p{\columnwidth}@{}}{
\centering\arraybackslash
\textit{Role of GDCF}} \\
\midrule
w/o DGL-Fusion
& 70.72 & 79.41 & 70.56 \\
Static Equal Ratio
& 70.93 & 82.35 & 71.47 \\
Static Global-heavy Ratio
& 72.03 & 82.35 & 71.81 \\
No Global Weight Constraint
& 71.93 & 80.88 & 71.99 \\
\midrule
\multicolumn{4}{@{}p{\columnwidth}@{}}{
\centering\arraybackslash
\textit{Role of Dynamic Modality-Specific MoEs}} \\
\midrule
w/o Dynamic MoEs
& 71.23 & 66.18 & 69.36 \\
w/o Cross-modal Guidance
& 70.12 & 58.82 & 72.21 \\
MLP Expert Replacement
& 70.32 & 79.41 & \textbf{72.28} \\
Random Routing
& 71.13 & 75.00 & 71.92 \\
Uniform Routing
& 72.54 & 77.94 & 71.87 \\
\midrule
\multicolumn{4}{@{}p{\columnwidth}@{}}{
\centering\arraybackslash
\textit{Global--Local Branch Controls}} \\
\midrule
Global Branch Only
& 72.33 & 79.41 & 72.01 \\
Local Branch Only
& 70.32 & 66.18 & 70.40 \\
\midrule
\multicolumn{4}{@{}p{\columnwidth}@{}}{
\centering\arraybackslash
\textit{Capacity Control}} \\
\midrule
Parameter-matched MLP
& 70.32 & 79.41 & \textbf{72.28} \\
\midrule
\multicolumn{4}{@{}p{\columnwidth}@{}}{
\centering\arraybackslash
\textit{Unimodal Controls}} \\
\midrule
Text Only
& 71.43 & 73.53 & 52.37 \\
Audio Only
& 54.43 & 47.06 & 67.03 \\
Vision Only
& 51.81 & 55.88 & 56.46 \\
\midrule
\textbf{DGC-Net (Full)}
& \textbf{74.14} & \textbf{83.82} & 72.16 \\
\bottomrule
\end{tabularx}
\end{table}

Table~\ref{tab:additional_ablation} reports runs with seeds
87, 102, and 20 for UR-FUNNY, MUStARD, and CHERMA, respectively.
Their full-model accuracies, $74.14\%$, $83.82\%$, and $72.16\%$,
are the references for these ablations. The first two differ from
the overall-comparison runs in
Table~\ref{tab:humor_sarcasm_results}.

\paragraph{Controlled variants.}
w/o DGL-Fusion substitutes direct MLP fusion. The static-ratio
variants fix interpolation weights, and No Global Weight
Constraint removes their bound. w/o Dynamic MoEs removes expert
selection and refinement; w/o Cross-modal Guidance jointly removes
RatioNet and auxiliary globals. MLP Expert Replacement changes
the expert transformation, while Random and Uniform Routing
replace learned routing. Branch-only, parameter-matched MLP,
and unimodal variants provide branch, capacity, and input controls.

\paragraph{Effects of Fusion Design.}
Replacing DGL-Fusion with direct MLP fusion reduces accuracy by
$3.42$, $4.41$, and $1.60$ percentage points on UR-FUNNY,
MUStARD, and CHERMA, respectively. Both fixed-ratio variants
also perform below the full model on all three datasets.
Removing the global weight constraint yields accuracies of
$71.93\%$, $80.88\%$, and $71.99\%$.
These comparisons support the evaluated dynamic compensation
design, although accuracy differences alone do not establish
that the learned coefficients measure representation reliability.

\paragraph{Effects of Expert Routing and Refinement.}
Removing the Dynamic Modality-Specific MoEs Module reduces
accuracy by $2.91$, $17.64$, and $2.80$ percentage points,
respectively. Removing cross-modal guidance reduces accuracy
by $4.02$ and $25.00$ percentage points on UR-FUNNY and
MUStARD, but slightly increases CHERMA accuracy from
$72.16\%$ to $72.21\%$.
Random and uniform routing underperform the full model on
all three datasets. Replacing the CLGSI Subnet Experts with
MLP experts reduces accuracy by $3.82$ and $4.41$ percentage
points on UR-FUNNY and MUStARD, while increasing CHERMA
accuracy to $72.28\%$. The benefits of cross-modal guidance
and the expert architecture therefore vary across tasks.

\paragraph{Branch and Capacity Comparisons.}
The full model exceeds both the global-only and local-only
controls on all three datasets. Against the parameter-matched
MLP, it improves accuracy by $3.82$ and $4.41$ percentage
points on UR-FUNNY and MUStARD, but trails by $0.12$
percentage points on CHERMA. These results support the
combination of global and local branches in the evaluated
runs, while showing that the proposed architecture does not
uniformly outperform a dense model with matched parameter count.

\paragraph{Unimodal Comparisons.}
The full model exceeds every unimodal variant. Text is the strongest
individual modality on UR-FUNNY and MUStARD, with accuracies of
$71.43\%$ and $73.53\%$, respectively, whereas audio is strongest
on CHERMA at $67.03\%$. These runs indicate task-dependent
unimodal performance and benefits from multimodal inputs. With one
seed per variant and dataset, statistical significance across
training runs remains unestablished.

\section{Case Study}

Table~\ref{tab:case_mosei} presents two correct predictions for
clear positive and negative expressions. Their absolute errors are 0.15 and 0.18, respectively,
showing close agreement with the annotated scores. Both examples have higher
global text weights than audio and visual weights. Case 2567 has a
negative label but a near-zero positive prediction for an utterance
containing both ``enjoy'' and ``did not.'' For this failed sample,
$y=-1.33$, $\hat{y}=0.05$, and $|y-\hat{y}|=1.38$, so the near-zero
output has the opposite polarity to its annotation. These examples
illustrate gating behavior and remaining errors without establishing
modality reliability.

\begin{table}[ht]
\caption{Selected CMU-MOSEI cases. $y$ and $\hat{y}$ denote
true and predicted scores. Error is $|y-\hat{y}|$.
Weights follow text/audio/vision order.}
\label{tab:case_mosei}
\centering
\small
\setlength{\tabcolsep}{0pt}
\renewcommand{\arraystretch}{1.1}

\begin{tabular}{
@{}
p{0.52\columnwidth}
>{\centering\arraybackslash}p{0.16\columnwidth}
>{\centering\arraybackslash}p{0.16\columnwidth}
>{\centering\arraybackslash}p{0.16\columnwidth}
@{}
}
\toprule
Global weights & $y$ & $\hat{y}$ & Error \\
\midrule

\multicolumn{4}{@{}p{\columnwidth}@{}}{
Case 3918: but it a great color I love
} \\
$(0.819,\,0.737,\,0.729)$
& 2.00 & 2.15 & 0.15 \\
\midrule

\multicolumn{4}{@{}p{\columnwidth}@{}}{
Case 2771: it just so stupid.
} \\
$(0.827,\,0.724,\,0.732)$
& -3.00 & -2.82 & 0.18 \\
\midrule

\multicolumn{4}{@{}p{\columnwidth}@{}}{
Case 2567: enjoy this one I did not it has the same
} \\
$(0.752,\,0.736,\,0.729)$
& -1.33 & 0.05 & 1.38 \\

\bottomrule
\end{tabular}
\end{table}
\paragraph{Interpretation of the global weights.}
The reported coefficients control the global--local interpolation
term within each modality. They are not a probability distribution
over modalities and therefore need not sum to one.
Cases 3918 and 2771 have similar global text weights
($0.819$ and $0.827$), although their sentiment labels have
opposite polarities. Thus, a larger global weight should not
be interpreted as a more positive prediction.
Case 2567 also satisfies the global-favoring constraint but
remains incorrectly predicted, showing that this constraint
alone does not ensure accurate sentiment interpretation.
% The final decision also depends on the encoded features,
% expert transformations, and subsequent fusion operations;
% the displayed weights alone cannot explain prediction correctness.

\paragraph{Local compensation in the selected cases.}
Within the interpolation branch, the local coefficient is
$1-\lambda_m$. The local text weights are therefore $0.181$,
$0.173$, and $0.248$ for Cases 3918, 2771, and 2567,
respectively. The failed case assigns a larger interpolation
weight to local text than either correctly predicted case.
Its local audio and vision weights are $(0.264,0.271)$,
compared with $(0.263,0.271)$ and $(0.276,0.268)$ for the
two correct cases. These values are derived directly from
the displayed global coefficients. They illustrate that the
balance can vary across samples, while a larger local share
does not necessarily accompany a correct prediction.
The coefficients apply to the interpolation term; residual
compensation and the attention-based fusion path also enter
the final representation. Consequently, these shares should
not be read as percentages of the final sentiment score
attributable to individual modalities.
These observations concern the selected examples and do not
establish a dataset-wide relation between local weighting and
prediction error. Assessing that relation would require aggregate
analysis over the complete test set.